# The Management of Context-Sensitive Features: A Review of Strategies

**Peter Turney**

Institute for Information Technology
National Research Council Canada
Ottawa, Ontario, Canada, K1A 0R6
peter@ai.iit.nrc.ca

## Abstract

In this paper, we review five heuristic strategies for handling context-sensitive features in supervised machine learning from examples. We discuss two methods for recovering lost (implicit) contextual information. We mention some evidence that hybrid strategies can have a synergetic effect. We then show how the work of several machine learning researchers fits into this framework. While we do not claim that these strategies exhaust the possibilities, it appears that the framework includes all of the techniques that can be found in the published literature on context-sensitive learning.

## 1  Introduction

This paper is concerned with the management of context for supervised machine learning from examples. We assume the standard machine learning framework, where examples are represented as vectors in a multidimensional feature space (also known as the *attribute-value* representation). We assume that a teacher has partitioned a set of training examples into a finite set of classes. It is the task of the machine learning system to induce a model for predicting the class of an example from its features.

In many learning tasks, we may distinguish three different types of features: *primary, contextual*, and *irrelevant* features (Turney, 1993a, 1993b). Primary features are useful for classification when considered in isolation, without regard for the other features. Contextual features are not useful in isolation, but can be useful when combined with other features. Irrelevant features are not useful for classification, either when considered alone or when combined with other features.

We believe that primary features are often context-sensitive. That is, they may be useful for classification when considered in isolation, but the learning algorithm may perform even better when we take the contextual features into account. This paper is a survey of strategies for taking contextual features into account. The paper is motivated by the belief that contextual features are pervasive. In support of this claim, Table 1 lists some of the examples of contextual features that have been examined in the machine learning literature. Many standard machine learning datasets (Murphy & Aha, 1996) contain contextual features, although this is rarely (explicitly) exploited. For example, in medical diagnosis problems, the patient's gender, age, and weight are often available. These features are contextual, since they (typically) do not influence the diagnosis when they are considered in isolation.

In Section 2, we list five heuristic strategies for managing context. We often neglect context, because of its very ubiquity; however, it is sometimes possible to recover hidden (implicit, missing) contextual information. Section 3 discusses two techniques (clustering and time sequence) for exposing hidden context. Section 4 reviews evidence that hybrid strategies can perform better than the sum of the component strategies (synergy). Section 5 briefly surveys the literature on context-sensitive learning and shows how the work of various researchers fits into the framework we present here. We conclude in Section 6.

Table 1: Some examples from the machine learning literature.

| Task | Primary Features | Contextual Features | Reference |
|---|---|---|---|
| image classification | local properties of the images | lighting conditions (bright, dark) | Katz *et al.* (1990) |
| speech recognition | sound spectrum information | speaker's accent (American versus British) | Pratt *et al.* (1991) |
| gas turbine engine diagnosis | thrust, temperature, pressure | weather conditions (temperature, humidity) | Turney & Halasz (1993), Turney (1993a, 1993b) |
| speech recognition | sound spectrum information | speaker's identity and gender | Turney (1993a, 1993b), Kubat (1996) |
| hepatitis prognosis | medical data | patient's age | Turney (1993b) |
| speech recognition | sound spectrum information | neighbouring phonemes | Watrous (1991) |
| speech recognition | sound spectrum information | speaker's identity | Watrous (1993) |
| heart disease diagnosis | electrocardiogram data | patient's identity | Watrous (1995) |
| tonal music harmonization | meter, tactus, local key | to be discovered by the learner | Widmer (1996) |

## 2 Strategies for Managing Context

Figure 1 illustrates our intuition about a common type of context-sensitivity. Let us consider a simple example: Suppose we are attempting to distinguish healthy people (class A) from sick people (class B), using an oral thermometer. Context 1 consists of temperature measurements made on people in the morning, after a good sleep. Context 2 consists of temperature measurements made on people after heavy exercise. Sick people tend to have higher temperatures than healthy people, but exercise also causes higher temperature. When the two contexts are considered separately, diagnosis is relatively simple. If we mix the contexts together, correct diagnosis becomes more difficult.

Katz *et al.* (1990) list four strategies for using contextual information when classifying. In earlier work (Turney, 1993a, 1993b), we named these strategies *contextual nor-*

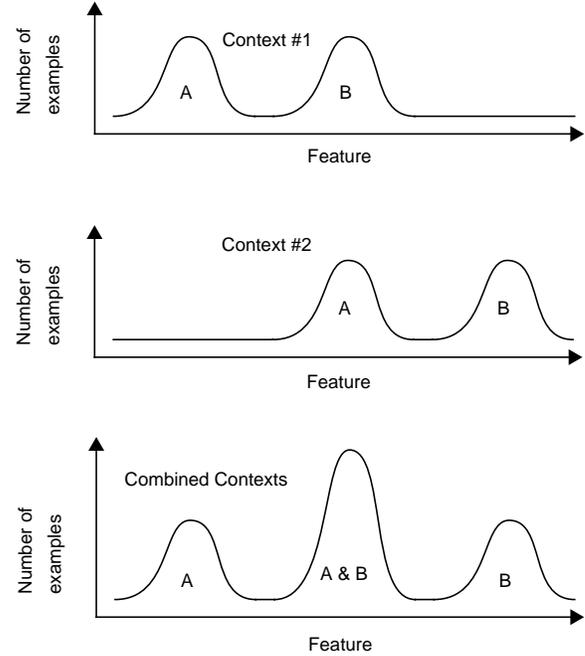

Figure 1. The result of combining samples from different contexts.

*malization, contextual expansion, contextual classifier selection,* and *contextual classification adjustment*.

**Strategy 1:** *Contextual normalization:* Contextual features can be used to normalize context-sensitive primary features, prior to classification. The intent is to process context-sensitive features in a way that reduces their sensitivity to context. For example, we may normalize each feature by subtracting the mean and dividing by the standard deviation, where the mean and deviation are calculated separately for each different context. See Figure 2.

**Strategy 2:** *Contextual expansion:* A feature space composed of primary features can be expanded with contextual features. The contextual features can be treated by the classifier in the same manner as the primary features. See Figure 3.

**Strategy 3:** *Contextual classifier selection:* Classification can proceed in two steps: First select a specialized classifier from a set of classifiers, based on the contextual features. Then apply the specialized classifier to the primary features. See Figure 4.

**Strategy 4:** *Contextual classification adjustment:* The two steps in contextual classifier selection can be reversed: First classify, using only the primary features. Then make an adjustment to the classification, based on the contextual features. The first step (classification using primary features alone) may be done by either a single classifier or multiple classifiers. For example, we might combine multiple specialized classifiers, each trained in a different context. See Figure 5.

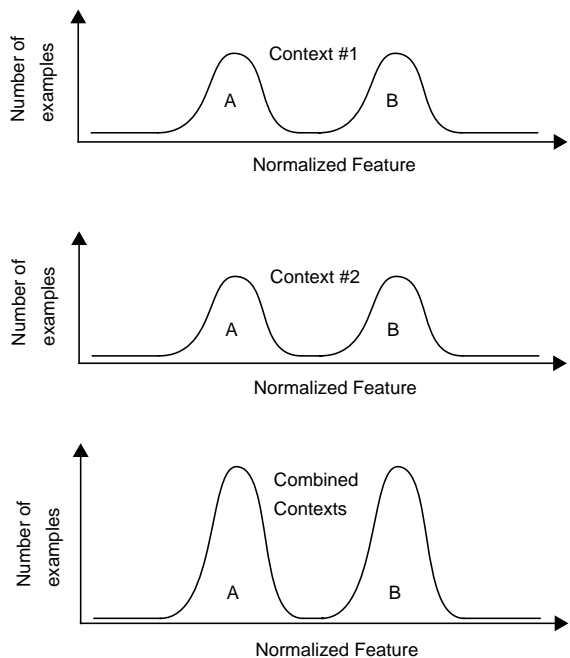

Figure 2. Contextual normalization: The result of combining normalized samples from different contexts.

In our previous work (Turney, 1993a, 1993b), we discussed a strategy that was not included in the list of four strategies given by Katz *et al.* (1990). We called this strategy *contextual weighting*.

**Strategy 5:** *Contextual weighting:* The contextual features can be used to weight the primary features, prior to classification. The intent of weighting is to assign more importance to features that, in a given context, are more useful for classification. Contextual selection of features (not to be confused with contextual selection of classifiers) may be viewed as an extreme form of contextual weighting: the selected features are considered important and the remaining features are ignored. See Figure 6.

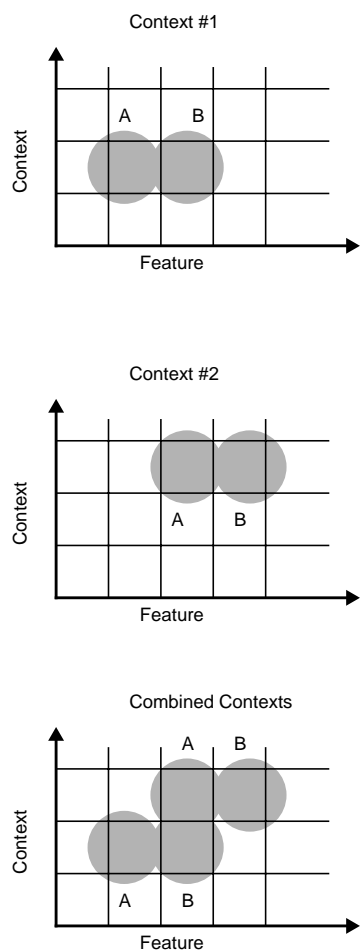

Figure 3. Contextual expansion: The result of combining expanded samples from different contexts.

## 3 Implicit Context

So far, we have been concerned with data in which contextual features are explicitly represented. Unfortunately, contextual information is often omitted from a dataset. Because we tend to take context for granted, we neglect to record the context of an observation. Fortunately, it is sometimes possible to recover contextual information. In this section, we consider two methods for recovering missing (hidden, implicit) contextual features. First, unsupervised clustering algorithms may be able to recover lost context (Aha, 1989; Aha & Goldstone, 1992; Domingos, 1996). Second, the temporal sequence of the instances may imply contextual information (Kubat, 1989; Widmer & Kubat, 1992, 1993, 1996).

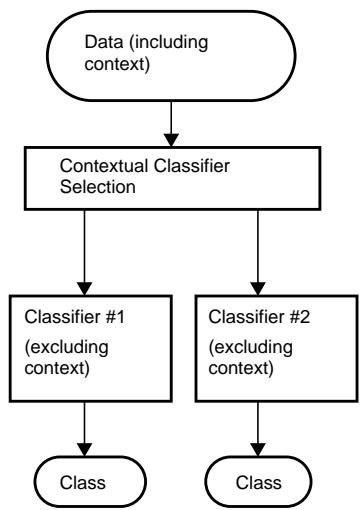
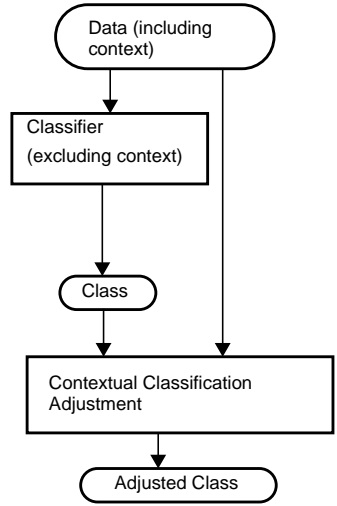
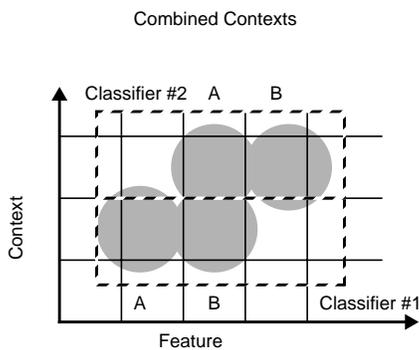

Figure 4. Contextual classifier selection: Different classifiers are used in different contexts.

We believe that clusters that are generated by unsupervised clustering algorithms typically capture shared context. That is, if two cases are assigned to the same cluster, then they likely share similar contexts. Therefore, if we cluster cases by their primary features, then members of the same cluster will tend to belong to the same class and the same context. More precisely, the likelihood that they belong to the same class and context is greater than the likelihood for the samples from the general population.

If we are given a dataset where there are only primary features, because the importance of contextual features was overlooked when the data were collected, we can use a clustering algorithm to recreate the missing contextual features. For example, we can label each case according to the cluster in which it belongs, and then we can introduce a new contextual feature of the form Cluster = Label. An alternative approach would be to integrate a form of clus-tering with a concept learning algorithm, instead of separating the clustering process from the classification process. This approach has been used by several researchers, with some success (Aha, 1989; Aha & Goldstone, 1992; Domingos, 1996).

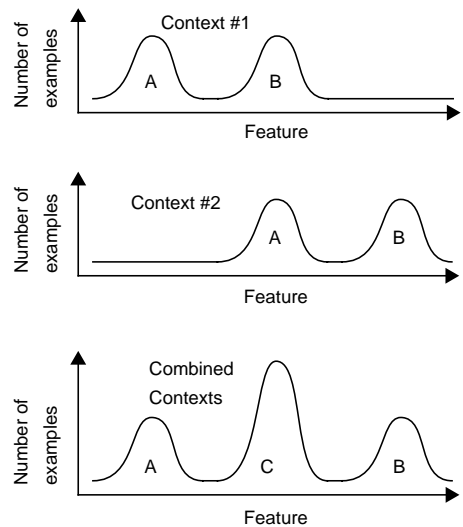

Figure 5. Contextual classification adjustment: The classification is adjusted for different contexts.

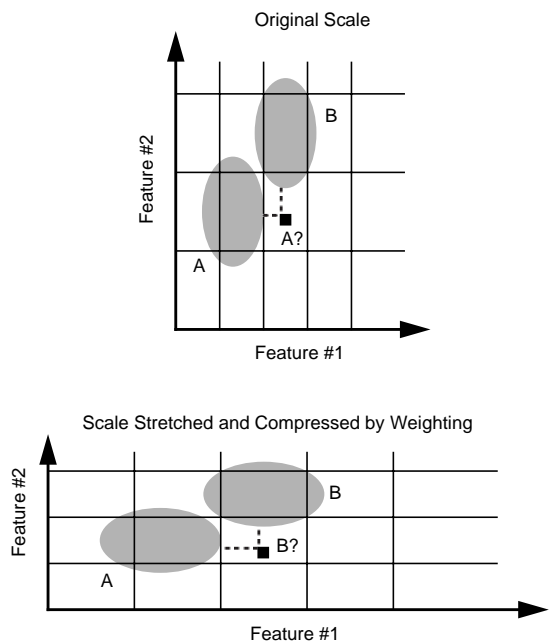

Figure 6. Contextual weighting: The impact of weighting on classification.

A feature of the form Cluster = Label might not be purely contextual, since clusters may be predictive of the class. Some of the success of approaches that combine clustering and classification may be due to this. Further research is required to determine whether clusters tend to be contextual or primary.

Another way to recover lost contextual information is to use temporal information, if it is available. We believe that events that occur close together in time tend to share context. If the records in a database contain a field for the date, this information might be used to expose hidden contextual information. We could introduce a new feature of the form Time = Date. Depending on what strategy we use for handling context, it may be useful to convert the time into a discrete feature.

In incremental learning, the order in which examples are encountered by the learner may correspond to the timing of the examples. In batch learning, the order of the examples in the file may correspond to the timing. We can introduce a new feature of the form Order = Number. Again, it may be useful to discretize this feature.

We believe that the FLORA algorithm (an incremental algorithm) is implicitly using the order of the examples to recover lost contextual information (Kubat, 1989; Widmer & Kubat, 1992, 1993, 1996). The FLORA algorithm is essentially an instance of the contextual classifier selection strategy (Strategy 3 in Section 2). The context is used to select the appropriate classifier from a set of possible classifiers. The interesting innovation is that the context is implied in the order of presentation of the examples.

## 4 Hybrid Strategies

Various combinations of the above strategies are possible. For example, we experimented with all eight possible combinations of three of the strategies (contextual normalization, contextual expansion, and contextual weighting) in two different domains, vowel recognition and hepatitis prognosis (Turney 1993a, 1993b).

In the vowel recognition task, the accuracy of a nearest-neighbour algorithm with no mechanism for handling context was 56%. With contextual normalization, contextual expansion, and contextual weighting, the accuracy of the nearest-neighbour algorithm was 66%. The sum of the improvement for the three strategies used separately was 3%, but the improvement for the three strategies together was 10% (Turney, 1993a, 1993b). There is a statistically significant synergetic effect in this domain.

In the hepatitis prognosis task, the accuracy of a nearest-neighbour algorithm with no mechanism for handling context was 71%. With contextual normalization, contextual expansion, and contextual weighting, the accuracy of the nearest-neighbour algorithm was 84%. The sum of the improvement for the three strategies used separately was 12%, but the improvement for the three strategies together was 13% (Turney, 1993b). The synergetic effect is not statistically significant in this domain.

One area for future research is to discover the circumstances under which there will be a synergy when strategies are combined. Another area for future research is to extend the experiments to all 32 possible combinations of the five strategies.

## 5 Applying the Framework to the Research Literature

The preceding sections of this paper have sketched a framework for categorizing strategies for learning in context-sensitive domains. We will now apply this scheme to a sample of the research literature. Table 2 shows how some of the papers fit into our structure. All of the papers we have read so far appear to be consistent with the framework.

Table 2: A classification of some of the literature on learning in context-sensitive domains.

| Reference | Context Management (Section 2) | Context Recovery (Section 3) |
|---|---|---|
| Aha (1989) | Weighting | Implicit — clustering |
| Aha and Goldstone (1992) | Weighting | Implicit — clustering |
| Bergadano *et al.* (1992) | Adjustment | Implicit — clustering |
| Domingos (1996) | Weighting | Implicit — clustering |
| Katz *et al.* (1990) | Selection | Explicit |
| Kubat (1996) | Selection, Adjustment | Explicit |
| Michalski (1987, 1989, 1990) | Adjustment | Implicit — clustering |
| Pratt *et al.* (1991) | Adjustment | Implicit — clustering |
| Turney (1993a, 1993b) | Normalization, Expansion, Weighting | Explicit |
| Turney and Halasz (1993) | Normalization | Explicit |
| Watrous (1991) | Adjustment | Explicit |
| Watrous (1993) | Normalization | Explicit |
| Watrous and Towell (1995) | Adjustment | Explicit |
| Widmer and Kubat (1992, 1993, 1996) | Selection | Implicit — temporal sequence |
| Widmer (1996) | Selection | Explicit |

In Table 2, *context management* refers to the five heuristics for managing context-sensitive features that are discussed in Section 2; *context recovery* refers to the method for recovering lost contextual features, as discussed in Section 3. *Explicit* means that the contextual features are explicitly present in the datasets. *Implicit* means that the contextual features were not recorded in the data, so the learning algorithm must attempt to recover lost contextual information. The implicit contextual information may be recovered either by clustering the data or exploiting the temporal sequence of the examples.

## 6 Conclusion

This paper briefly surveyed the literature on machine learning in context-sensitive domains. We found that there are five basic strategies for managing context-sensitive features and two strategies for recovering lost context. Combining strategies appears to be beneficial.

A survey such as this is the first step towards a scientific treatment of context-sensitive learning. Many open questions are raised: Is the list of strategies complete? Can the strategies be formally justified? What is the explanation of the synergy effect? These are topics for further research.

## Acknowledgments

Thanks to Miroslav Kubat and Peter Clark for sharing their ideas about context in numerous discussions with me. Thanks to Joel Martin and Dale Schuurmans for their comments on an earlier version of this paper. Thanks to two anonymous referees of the *Workshop on Learning in Context-Sensitive Domains* for their comments on an earlier version of this paper. Thanks to Miroslav Kubat for providing additional references.